\documentclass[screen]{acmart}
\usepackage{url}
\usepackage{subcaption}
\usepackage{tabularx}
\usepackage{multicol}
\usepackage{multirow}
\usepackage{algorithm}
\usepackage{graphicx}%
\usepackage{amsthm}%
\usepackage{algorithmicx}%
\usepackage{algpseudocode}%
\AtBeginDocument{%
  \providecommand\BibTeX{{%
    \normalfont B\kern-0.5em{\scshape i\kern-0.25em b}\kern-0.8em\TeX}}}

\copyrightyear{2024}
\acmYear{2024}
\setcopyright{rightsretained}

\acmConference[NLPIR 2024]{2024 8th International Conference on Natural Language Processing and Information Retrieval}{December 13--15, 2024}{Okayama, Japan}
\acmBooktitle{2024 8th International Conference on Natural Language Processing and Information Retrieval (NLPIR 2024), December 13--15, 2024, Okayama, Japan}\acmDOI{10.1145/3711542.3711594}
\acmISBN{979-8-4007-1738-3/24/12}

\begin{document}

\title{Data Augmentation With Back translation for Low Resource languages: A case of English and Luganda}

\author{Richard Kimera}
\email{kimirchies@handong.ac.kr}
\affiliation{%
  \institution{Mbarara University of Science and Technology, Dept. of Information Technology}
  \streetaddress{Mbarara-Kabale Highway}
  \city{Mbarara}
  \state{}
  \country{Uganda}
  \postcode{256}
}

\author{DongNyeong Heo}
\affiliation{%
  \institution{Handong Global University, Dept. of Computer Science and Electrical Engineering}
  \streetaddress{Handong-Ro, 558}
  \city{Pohang}
  \state{Gyeongbuk}
  \country{South Korea}
  \postcode{37554}
}
\email{sjglsks@gmail.com}

\author{Daniela N. Rim}
%%\authornotemark[2]
\affiliation{%
  \institution{Handong Global University, Dept. of Computer Science and Electrical Engineering}
  \streetaddress{Handong-Ro, 558}
  \city{Pohang}
  \state{Gyeongbuk}
  \country{South Korea}
  \postcode{37554}
}
\email{danielarim@handong.ac.kr}

\author{Heeyoul Choi}
%%\authornotemark[3]
\affiliation{%
  \institution{Handong Global University, Dept. of Computer Science and Electrical Engineering}
  \streetaddress{Handong-Ro, 558}
  \city{Pohang}
  \state{Gyeongbuk}
  \country{South Korea}
  \postcode{37554}}
\email{hchoi@handong.edu}

\renewcommand{\shortauthors}{Richard, DongNyeong, Daniela and Choi}

\begin{abstract}
In this paper, we explore the application of Back translation (BT) as a semi-supervised technique to enhance Neural Machine Translation (NMT) models for the English-Luganda language pair, specifically addressing the challenges faced by low-resource languages. The purpose of our study is to demonstrate how BT can mitigate the scarcity of bilingual data by generating synthetic data from monolingual corpora. Our methodology involves developing custom NMT models using both publicly available and web-crawled data, and applying Iterative and Incremental Back translation techniques. We strategically select datasets for incremental back translation across multiple small datasets, which is a novel element of our approach. The results of our study show significant improvements, with translation performance for the English-Luganda pair exceeding previous benchmarks by more than 10 BLEU score units across all translation directions. Additionally, our evaluation incorporates comprehensive assessment metrics such as SacreBLEU, ChrF2, and TER, providing a nuanced understanding of translation quality. The conclusion drawn from our research confirms the efficacy of BT when strategically curated datasets are utilized, establishing new performance benchmarks and demonstrating the potential of BT in enhancing NMT models for low-resource languages.
\end{abstract}



%%\received{20 February 2007}
%%\received[revised]{12 March 2009}
%%\received[accepted]{5 June 2009}

\maketitle

\section{Introduction}
Data scarcity presents a profound challenge in training Neural Machine Translation (NMT) models, especially for low resource languages (LRLs) \cite{sun2021machine, cheng2019joint} for instance, Luganda, whose online presence is also limited. This challenge primarily arises from the reliance of most NMT models on extensive bilingual corpora which consists of datasets containing pairs of equivalent text segments in two languages \cite{khenglawtaddressing}. Typically curated by human experts, often linguists with specialized knowledge of the source and target languages, the creation of such corpora is a time-intensive process. The associated costs make the compilation of these datasets prohibitive, particularly for multilingual communities in low-resource settings \cite{gatiatullin2022toolset}. This presents a formidable barrier to creating comprehensive datasets that encompass all indigenous languages, thereby hampering the development and proliferation of NMT-powered systems and further widening the digital divide.

In the context of this research, ``low resourcefulness" specifically denotes a marked shortage or outright absence of digital text data necessary for developing robust and reliable NMT models. This contrasts with high-resource languages (HRLs), which enjoy a wealth of data, including extensive digital texts (bilingual and monolingual data), and pre-trained language models, enabling the creation of sophisticated, and good translation systems that leverage advanced machine-learning techniques.

Such disparity in resource availability fosters a technological inequity, with HRLs advancing in digital presence while LRLs languish, underrepresented, and underserved in the digital domain. This inequity limits access to information and technology for speakers of LRLs and may even threaten these languages' survival in the digital epoch.

In this paper, we will use Luganda as a representative language for LRLs and English as a HRL. Luganda, is a language from Uganda, an East African nation with a multilingual tapestry. Here, NMT systems have the potential to significantly enhance translation and communication. The country has two official imported languages, i.e. English and Swahili. Despite English's prevalence and role as the primary language and Swahili's regional prominence, local communication channels often broadcast in a plethora of native languages. Luganda, spoken by over 20 million people in Uganda and neighboring countries, is a prime example of such languages \footnote{\url{https://blog.google/products/translate/24-new-languages/}}. As a morphologically rich language, Luganda features ten noun classes critical for sentence construction and exhibits a phonemic distinction that can convey starkly different meanings, as in ``\textit{kuta}" (meaning ``to release") versus ``\textit{Kutta}" (meaning ``to kill") \cite{baertlein2014luganda}. Its agglutinative nature allows for complex word constructions using prefixes, suffixes, and infixes, positioning Luganda as an exemplar among Bantu languages for NMT tasks \cite{ssentumbwe2019english}. This language faces low resourcefulness, just like many in sub-saharan Africa \cite{mukiibi2022makerere}.

One of the approaches used in NMT to address data scarcity is data augmentation methodologies, aimed at the generation of synthetic data \cite{oh2023data} and have been growing steadily. One such techniques is a semi-supervised approach called Back translation (BT). This standard technique involves training a model using bilingual data, it then uses the model to generate synthetic data by translating monolingual data. Finally, we merge the synthetically generated data with the bilingual data, which we use to update the standard model \cite{sennrich2015improving}. Effectively applied, BT can substantially enhance translation accuracy. BT has been utilized widely and proven to show improvements in NMT models \cite{sugiyama2019data, jin2022admix, aji2020fully}. It has also been used in LRL's for instance in South African languages \cite{mojapelo2023data} and in simulated environments where a small portion of a high resource language is used as a representation for data scarcity \cite{lamar2023measuring}.

Even though machine translation has been utilised on Luganda and English language pair \cite{kimera2022building, babirye2022building, rim2023mini, ssentumbwe2019english}, only a couple of works have applied BT. They focused on the use of standard BT \cite{sennrich2015improving}, for instance \cite{akera2022machine} and \cite{bapna2022building} utilized BT by finetuning multilingual models on custom built monolingual datasets. The application and training details provided could be enhanced and boosted as they do not explore how datasets from various sources could be combined while performing BT. Furthermore, the monolingual datasets used for either of these works is not publicly available, which escalates the data scarcity problem, motivating the need to build a publicly available dataset. Provision of both the public and monolingual datasets will aid reproducability and comparison of performance for future research, a component that is not possible today.     

In this paper, we begin by building both bilingual and monolingual datasets from previous works, we then crawl more monolingual data for the news domain, and then apply our BT approach. Hence, the work unfolds in three phases: 
\begin{enumerate}
    \item Curating and building a bilingual and monolingual dataset from existing open-source repositories;
    \item Employing web crawling and cleaning techniques to extract data from web sources.
    \item Introducing a back translation approach that continuously refines and improves the translation models by focusing training efforts on the most promising monolingual datasets. 
\end{enumerate}

Throughout our experiments, we achieve new performance benchmarks on English-Luganda translation system using the transformer architecture \cite{vaswani2017attention} and markedly improve the BLEU score between the bilingual and the BT models, by more than 10 points for both Lug2Eng and Eng2Lug translation directions. We also provide additional metrics, more effective for morphologically rich languages like Luganda, to demonstrate the nuanced performance of our models beyond what the BLEU score alone can reveal. The ChrF(character $n$-gram F-score) can better capture morphological similarities, where word inflections are crucial \cite{bapna2022building}. We provide the ChrF2 score to focus more on the quality of translation completeness. We further provide SacreBLEU as a standardized BLEU score, and Translation Error Rate (TER) so as to provide complementary information about the translation quality and as a measures of the number of edits required to change the hypothesis into a reference. These metrics can better capture morphological and grammatical correctness, fluency, and overall adequacy of the translations as seen in other low resource morphological languages such as Tamil \cite{hans2016improving}. 

\section{Related Works}
\subsection{NMT for Luganda and English}\label{subsec1}
 A bilingual dataset of 40,000 was built from Uganda conversations, storybooks, newspapers and online wikipedia articles, translated into multiple languages spoken in Uganda (Luganda, Acholi, Lumasaba and Runyankore) and used to build NMT models through fine-tuning the OPUS-MT model \cite{tiedemann2020opus} for both Eng2Lug and Lug2Eng \cite{babirye2022building}. Another research focused on building a bilingual dataset and training NMT models for Eng2Lug and Lug2Eng by applying the transformer model \cite{vaswani2017attention} and performing a hyperparameter search by utilising publicly available datasets \cite{kimera2022building}. Unlike \cite{babirye2022building} that was custom built from various online and offline sources in Uganda for example newspapers, wikipedia, and storybooks, \cite{kimera2022building} merged three publicly available datasets totaling to 41,070 parallel sentences. Two of the datasets were downloaded from the zenodo web portal, the text used was collected from Luganda teachers, students, and freelancers\footnote{https://zenodo.org/}. The third dataset was from sunbird AI which they built from social media, transcripts from radio, online newspapers, articles, blogs, text contributions from Makerere University NLP community, and farmer responses from surveys \cite{akera2022machine}. 
 
 Another approach focused on improving training efficiency through data sorting. Here sentences with similar length were randomly sampled from a dataset and Lug2Eng and Eng2Lug models were trained on mini-batches \cite{rim2023mini}, using the same dataset as \cite{kimera2022building}. Other researchers utilized the Bible dataset and performed Statistical Machine Translation (SMT) through the segmentation of Luganda nouns based on their classes. This research focused on the morphological structure of Luganda \cite{ssentumbwe2019english}. Our research builds upon the bilingual datasets from \cite{kimera2022building} and the monolingual dataset from \cite{babirye2022building}. We further expand on the size of these datasets' since it is very crucial for a better NMT system. %We also clean the collected datasets to make them more suitable for building the translation models. 

\subsection{Back translation for Luganda and English}\label{subsec2}
Even though BT has been applied in LRLs \cite{lamar2023measuring, bapna2022building, gitau2023textual}, its application to Luganda and Ugandan languages is still growing. A major attempt to using BT on Luganda focused on fine-tuning the OPUS-MT model using a custom-built dataset, and collecting monolingual data from local sources and applying BT. The techniques yielded an improvement in the BLEU scores by approximately +3 units in either Eng2Lug or Lug2Eng directions \cite{akera2022machine}. Another major contribution is done by Google among other LRLs including Luganda. They used BT to train NMT models for Eng2Lug and Lug2Eng. %achieve SpBLEU scores of 17.3 for Eng2Lug and 29.3 for Lug2Eng.
They provide both the BLEU score and ChRF score for Eng2Lug to show the importance of other metrics on morphological languages. They used a custom dataset as well web crawling of 2 million Luganda monolingual data \cite{bapna2022building}. The monolingual datasets for \cite{akera2022machine} and \cite{bapna2022building} are not publicly available hence the data scarcity problem became one of the challenges we sought to address and also encouraged us to build a publicly available dataset. We also noticed that there was a need to provide more evaluation metrics since the BLEU score alone is not a perfect measure, and additional metrics could be better for fluency \cite{akera2022machine, bapna2022building}. 

% Furthermore, we employ various BT techniques to enhance translation quality. Given the scarcity of detailed methodologies applying BT to Luganda, we explore iterative BT, incremental BT, and develop our own BT approach. Our BT method is inspired by the need to select the most suitable dataset from multiple datasets collected from a variety of sources. Our BT then focuses on selecting the optimal monolingual and translation models to train better BT models. We conduct experiments on monolingual data from multiple open-source platforms, as well as web-crawled data, which we use to augment existing datasets.   

Furthermore, we enhance translation quality using various BT techniques, including iterative and incremental BT, and introduce a custom BT approach tailored for optimal dataset selection from diverse sources. By experimenting with monolingual data from open-source platforms and web-crawled sources, we augment existing datasets to improve model performance.

\section{Methods}
In this section, we describe the construction of both the bilingual and monolingual datasets. Subsequently, we discuss the application of well-known BT approaches. We further detail how our specific approach was implemented and the improvements it offers and conclude it by the metrics used.

\subsection{Data collection}
We began by collecting publicly available data, which included both bilingual and monolingual datasets in English and Luganda. Most datasets were sourced from Uganda, thereby reflecting the Ugandan context. To augment our collection of monolingual data, we gathered additional texts from various news sources, including websites and YouTube news channels.

\subsubsection{Combined bilingual datasets} \label{subsubsec1}
The datasets were obtained from open-source directories and subsequently merged to create a unified dataset for the experiments. Table \ref{table:bilingual} presents the names of the datasets, the total number of sentences in each, and their sources of download. 
    
    \begin{table}[ht]
    \centering
    \caption{English and Luganda Bilingual Data Source}
    \label{table:bilingual}
    %\begin{tabular}{|m{4cm}|m{1.5cm}|m{1.5cm}|}
    %\begin{tabular}{|c|c|c|}
    \begin{tabularx}{\textwidth}{|X|X|X|}
    \hline
    \textbf{Name} & \textbf{Total sentences} & \textbf{source} \\ \hline
    NMT for English and Luganda & 41,070 & \cite{kimera2022building} \\ \hline
    Translation Initiative for Covid-19 & 15,022 & \cite{anastasopoulos2020tico} \\ \hline
    Makerere Sentiment Corpus & 10,000 & \cite{DVN/XSGIKR_2023} \\ \hline
    Bible WorldProject & 32,291 & Webscraped\footnotemark[2] 
    \\ \hline
    \end{tabularx}
    \end{table}
    \footnotetext[2]{https://www.wordproject.org/bibles/index.htm}

\subsubsection{Monolingual data}
The English data was collected from various sources as seen in Table \ref{table:mono_english} so does Luganda in Table \ref{table:mono_luganda}. 

All the web scraped data was extracted using the selenium API\cite{gundecha2015selenium}. The text was then cleaned by removing hyperlinks, repetitive text, code-mixed text, empty spaces, and special characters. 
    
    \begin{table}[ht]
    \centering
    \caption{Monolingual English Data Sources}
    \label{table:mono_english}
    %\begin{tabular}{|m{4.25cm}|m{1.3cm}|m{1.7cm}|}
    %\begin{tabular}{|c|c|c|}
    \begin{tabularx}{\textwidth}{|X|c|c|}
    \hline
    \textbf{Name} & \textbf{Total} & \textbf{Source} \\ \hline
    Digital Umuganda & 14,400 & Huggingface\footnotemark[3] \\ \hline
    Gamayun Mini kit & 5,000 & Gamayun\footnotemark[4]\\ \hline
    Tatoeba & 1,000,000 & Tatoeba\footnotemark[5]\\ \hline
    Youtube news headlines (NTV, NBS) + websites (Softpower and Monitor) & 20,005 & Webscraped\footnotemark[6]\\ \hline
    Softpower news Articles & 49,945 & Webscraped\footnotemark[7] \\ \hline
    Chimpreports news Articles & 90,284 & Webscraped\footnotemark[8]\\ \hline
    \end{tabularx}
    \end{table}
        \footnotetext[3]{https://huggingface.co/DigitalUmuganda}
        \footnotetext[4]{https://gamayun.translatorswb.org/download/gamayun-5k-english-hausa}
        \footnotetext[5]{https://tatoeba.org/en/downloads}
        \footnotetext[6]{https://www.youtube.com}
        \footnotetext[7]{https://softpower.ug/}
        \footnotetext[8]{https://chimpreports.com/category/news/}

    \begin{table}[h]
    \centering
    \caption{Monolingual Luganda Data Sources}
    \label{table:mono_luganda}
    %\begin{tabularx}{|m{4.25cm}|m{1.3cm}|m{1.7cm}|}
    %\begin{tabular*}{\textwidth}{@{\extracolsep\fill}|c|c|c|}
    \begin{tabularx}{\textwidth}{|X|c|c|}
    \hline
    \textbf{Name} & \textbf{Total} & \textbf{Source} \\ \hline
    Mozilla common voice & 128,491 & Huggingface\footnotemark[9]\\ \hline
    cc-100 (Web Crawl) & 78,556 & \cite{conneau2019unsupervised, wenzek2019ccnet} \\ \hline
    Masakhane & 9,809 & \cite{adelani2023masakhanews} \\ \hline
    Makerere Radio Dataset & 11,550 & \cite{mukiibi2022makerere} \\ \hline
    MakerereText and Speech & 100,000 & \cite{DVN/EQOWTW_2023} \\ \hline
    Youtube news headlines (NTV news (akawungeezi), NBS news (Amasengejje), Bukedde TV (Agataliiko nfuufu) and CBS FM website & 5,253 & Webscraped\footnotemark[10] \\ \hline
    \end{tabularx}
    \end{table}
        \footnotetext[9]{https://huggingface.co/mozilla-foundation}
        \footnotetext[10]{https://www.youtube.com}

\subsection{Bilingual NMT and back translation approaches}
 The bilingual dataset was split into training, testing and validation sets. We refer to the testing and validation splits as "default (T/V)" as depicted in Table \ref{table:bleu_default_e2l} and Table \ref{table:bleu_default_l2e}. We performed BT both with and without the Bible text. The reason why we use the Bible text was to show that while this text can reduce the data scarcity problem since it is widely available, its context may not be close to natural conversation, for instance the most widely available Bible text i.e the Bible WorldProject that was used in these experiments is the King James version, it contains words such as ``thee", ``thy", ``yea", ``believeth", ``strengtheneth", among others. 
 
 For all our BT experiments, the text was shuffled before model training. We also maintained the same BPE \cite{sennrich2015neural} code for bilingual and BT models. During the translation of the monolingual data, we used beam search to generate the synthetic data \cite{edunov2018understanding}.

 Because there was an imbalanced amount of data between the sources of bilingual dataset, the validation and testing sets could be biased to the majority. To mitigate this bias, we additionally expanded them with more general translation pairs, such as news and named it, ``newtest" as seen in Table \ref{table:eng_to_lug_newtest} and Table \ref{table:lug_to_eng_newtest}.

 \subsubsection{Iterative and Incremental BT}
 While performing BT, if the standard BT is repeated a couple of times until convergence is achieved, it is referred to as iterative BT (IteBT)\cite{hoang2018iterative}, and if monolingual data is increased in portions, it can be referred to as incremental BT (IncBT)\cite{hoang2018iterative}. We apply these two approaches during our experiments.

 \subsubsection{Our approach BT (OurBT)}
 At the core of OurBT as in Algorithm \ref{alg:OurBT} is the initialization of two primary translation models, denoted as $M_{EL}$ for English-to-Luganda and $M_{LE}$ for Luganda-to-English, trained on dataset $\mathcal{D}$, comprised of bilingual sentences $\mathbf{e}j$ and $\mathbf{l}j$. The iterative phase of the algorithm unfolds as a sequence of translation and retraining cycles over $N$ monolingual datasets, where each dataset is indexed by $i$. For each cycle, the algorithm employs $M_{LE}$ to translate the $i$-th set of monolingual Luganda sentences, $\mathcal{M}L^{(i)}$, yielding a synthetic English dataset $\mathcal{E}{synth}$, while concurrently, $M_{EL}$ translates the $i$-th set of monolingual English sentences, $\mathcal{M}E^{(i)}$, to generate a synthetic Luganda dataset $\mathcal{L}{synth}$. These synthetic datasets are used for further training of the models, producing new iterations $M_{EL}^{(i)}$ and $M_{LE}^{(i)}$. Based on the BLEU score, OurBT identifies the most relevant datasets $\mathcal{L}_{best}$ and $\mathcal{E}_{best}$, and models $M_{LE}^{best}$ and $M_{EL}^{best}$. Finally, the best monolingual datasets are translated using the selected models, followed by retraining the models on the resulting synthetic pairs, resulting in refined models $M_{EL}^*$ and $M_{LE}^*$. 
 
\begin{algorithm}
\caption{OurBT Algorithm}
\label{alg:OurBT}
\begin{algorithmic}[1]
\State \textbf{Input}: Bilingual dataset $\mathcal{D} = \{(\mathbf{e}_j, \mathbf{l}_j)\}$ with monolingual datasets $\mathcal{M}_E$ and $\mathcal{M}_L$ for English and Luganda respectively, and initialized NMT models $M_{EL}$ and $M_{LE}$.
\State \textbf{Output}: Enhanced NMT models $M_{EL}^*$ and $M_{LE}^*$

\State $M_{EL}, M_{LE} \gets \textsc{Train}(\mathcal{D})$

\Comment{Begin back-translation with monolingual datasets}
\For{$i \in \{1, \ldots, N\}$}
    \State $\mathcal{E}_{synth} \gets \textsc{Translate}(\mathcal{M}_L^{(i)}, M_{LE})$
    \State $\mathcal{L}_{synth} \gets \textsc{Translate}(\mathcal{M}_E^{(i)}, M_{EL})$
    
    \State $M_{EL}^{(i)} \gets \textsc{Train}(\mathcal{M}_E^{(i)}, \mathcal{L}_{synth})$
    \State $M_{LE}^{(i)} \gets \textsc{Train}(\mathcal{M}_L^{(i)}, \mathcal{E}_{synth})$
\EndFor

\Comment{Select models and datasets with the highest BLEU scores for final training}
\State $\mathcal{L}_{best}, \mathcal{E}_{best} \gets \textsc{SelecBasedOnBLEU}(\mathcal{M}_L, \mathcal{M}_E, M_{LE}, M_{EL})$
\State $M_{LE}^{best}, M_{EL}^{best} \gets \textsc{SelectBestBLEU}(M_{LE}, M_{EL})$

\State $\mathcal{E}_{final} \gets \textsc{Translate}(\mathcal{L}_{best}, M_{LE}^{best})$
\State $\mathcal{L}_{final} \gets \textsc{Translate}(\mathcal{E}_{best}, M_{EL}^{best})$

\State $M_{EL}^* \gets \textsc{Train}(\mathcal{L}_{best}, \mathcal{E}_{final})$
\State $M_{LE}^* \gets \textsc{Train}(\mathcal{E}_{best}, \mathcal{L}_{final})$

\State \textbf{Return} $M_{EL}^*$, $M_{LE}^*$
\end{algorithmic}
\end{algorithm}

\subsection{Evaluation metrics}
As emphasized earlier, the use of BLEU score alone as a metric to measure translation quality hurts Morphologically Rich languages. In their research that saw the addition of 27 low-resource languages to Google translate platform, Google demonstrated better performance with the use of ChrF score as compared to BLEU score for inflicted languages such as Kalaallisut and Luganda\cite{bapna2022building}. For our research, we considered the ChrF2 score, SacreBLEU and Translation Edit Rate (TER).

\section{Experiments}
\subsection{Experimental details}
 During training and validation, the inputs are processed in mini-batches of size 1000. Adam\cite{jais2019adam} was used as the optimizer for all models. If the validation loss plateaus or doesn't improve for 40 epochs, early stopping is enacted. For software, Python 3.8.10 was used, with support from the PyTorch library. For hardware, a Ubuntu server with 2 GPUs of type NVIDIA GeForce RTX 3090 is utilized for all experimentation purposes.

 Throughout our experiments, we applied Byte-Pair Encoding (BPE)\cite{sennrich2016neural}, and set the vocabulary to 10,000 subword units. BPE operates by dividing words into smaller subunits or subwords based on their occurrence frequency. This technique significantly enhances the model's ability to manage rare and unknown words by decomposing them into recognizable subunits.
 
 \subsubsection{Bilingual (BiT) and BT approaches}  
 In this stage, we carried out three experiments using the Transformer base model configuration as presented in the paper\cite{vaswani2017attention}, consisting of \textit{N=6,  $d_{\text{model}}$=512, $d_{\text{ff}}$=2048}, h=8, $d_{\text{k}}$=64, $d_{\text{v}}$=64, and $P_{\text{drop}}$=0.1.
 
 \begin{itemize}
    \item The first experiments were carried out on a dataset with Bible text, aimed at having a large data size as well as expanding the diversity of our dataset. IncBT was then applied. 
    \item In the second experiment, the Bible text was removed to have a more context-aware dataset, of which IncBT was applied. 
    \item The third experiment used the training dataset from the second experiment above, and the expanded testing and validation dataset. This was done to have a more general testset focused on news from the local context that was deemed to fit the web-scraped data. OurBT and IteBT were applied on this dataset.
 \end{itemize}

\subsection{Default testing and validation (T\&V) sets}
In all our results, we use bilingual to refer to the standard or base NMT model before applying BT. In Table \ref{table:bleu_default_e2l}, removal of the bible text led to an improved BLEU score. This was due to the use of King James Bible version whose context was different from the rest of the text. We applied IncBT on either datasets, first with the Bible text and second without the Bible text. The best dataset combination (YouTube news headlines) had a marginal improvement of the BLEU score by +0.41(39.46) and +0.11(53.88) for Eng2lug. This was because they were collected from Ugandan news text, hence matching the data context. For Lug2Eng, the BLEU scores did not yield any improvements but rather reduced. Table \ref{table:bleu_default_l2e} depicts this result.  

    \begin{table}[ht]
    \centering
    \caption{English to Luganda BLEU Scores (default T\&V)}
    \label{table:bleu_default_e2l}
    %\begin{tabular}{|m{0.5cm}|m{4cm}|m{2cm}|}
    %\begin{tabular}{|c|c|c|}
    \begin{tabularx}{\textwidth}{|c|X|c|}
    \hline
    \textbf{No} & \textbf{Dataset} & \textbf{BLEU Score} \\ \hline
    1 & Bilingual + Bible text & 39.05 \\ \hline
    2 & (1)+ Incremental BT & \textbf{39.46} \\ \hline
    3 & Bilingual - Bible text & 53.77 \\ \hline
    4 & (3)+ Incremental BT & \textbf{53.88} \\ \hline
    \end{tabularx}
    %\footnotetext{Source: This is an example of table footnote. This is an example of table footnote.}
    \end{table}
    %\end{tabular}
    %\end{table}

    \begin{table}[ht]
    \centering
    \caption{Luganda to English BLEU Scores (default T\&V)}
    \label{table:bleu_default_l2e}
    %\begin{tabular}{|m{0.5cm}|m{4cm}|m{2cm}|}
    %\begin{tabular}{|c|c|c|}
    \begin{tabularx}{\textwidth}{|c|X|c|}
    \hline
    \textbf{No} & \textbf{Dataset} & \textbf{BLEU Score} \\ \hline
    1 & Bilingual + Bible text & 44.36 \\ \hline
    2 & (1)+ Incremental BT & 35.65 \\ \hline
    3 & Bilingual - Bible text & 57.46 \\ \hline
    4 & (3)+ Incremental BT & 57.25 \\ \hline
    \end{tabularx}
    \end{table}

\subsection{Newtest testing and validation sets}
Using the newtest dataset and applying our approach (OurBT) improved the model performance. For Eng2Lug (Table \ref{table:eng_to_lug_newtest}), the BLUE score increased by 6.27 units. When we applied IteBT for 4 iterations until convergence, we further improved it to 10.58, with an increase in each iteration. The SacreBLEU and ChrF2 scores were higher than the BLEU scores due to Luganda being a morphologically rich language. 

    \begin{table}[ht]
    \centering
    \caption{English to Luganda Scores (newtest)}
    \label{table:eng_to_lug_newtest}
    %\begin{tabular}{|m{1.8cm}|m{0.8cm}|m{0.8cm}|m{1.4cm}|m{0.8cm}|m{0.8cm}|}
    %\begin{tabular}{|c|c|c|c|c|c|}
    \begin{tabularx}{\textwidth}{|X|X|X|X|X|X|}
    \hline
    \textbf{Model} & \textbf{BLEU} & \textbf{Gain} & \textbf{SacreBLEU} & \textbf{chrF2} & \textbf{TER} \\ \hline
    Bilingual & 29.67 & & 28.3 & 47.9 & 69.7 \\ \hline
    StandardBT & 32.29 & 2.49 & 31.2 & 50.2 & 70.2 \\ \hline
    OurBT & \textbf{35.94} & 6.27 & 34 & 51.9 & 67.2 \\ \hline
    iteration 1 & 37.96 & 8.29 & 36.1 & 52.6 & 65.7 \\ \hline
    iteration 2 & 39.31 & 9.64 & 37.3 & 53.1 & 65.9 \\ \hline
    iteration 3 & \textbf{40.25} & \textbf{10.58} & 38.2 & 53.2 & 65.4 \\ \hline
    \end{tabularx}
    \end{table}

OurBT improved the BLEU score for Lug2Eng model by 7.05, and applying iteBT continuously improved the performance up to 11.33 points in the same number of iterations as the Eng2Lug BT.   

    \begin{table}[ht]
    \centering
    \caption{Luganda to English Scores (newtest)}
    \label{table:lug_to_eng_newtest}
    %\begin{tabular}{|c|c|c||c|c|c|}
    \begin{tabularx}{\textwidth}{|X|X|X|X|X|X|}
    \hline
    \textbf{Model} & \textbf{BLEU} & \textbf{Gain} & \textbf{SacreBLEU} & \textbf{chrF2} & \textbf{TER} \\ \hline
    Bilingual & 32.92 & & 32.8 & 46.8 & 64.7 \\ \hline
    StandardBT & 33.37 & 0.45 & 32.8 & 49.4 & 68.6 \\ \hline
    OurBT & \textbf{39.97} & 7.05 & 38.4 & 52.3 & 63.4 \\ \hline
    iteration 1 & 41.94 & 9.02 & 40.9 & 52.9 & 63.2 \\ \hline
    iteration 2 & 43.42 & 10.5 & 43 & 53.4 & 61 \\ \hline
    iteration 3 & \textbf{44.25} & \textbf{11.33} & 44 & 53.8 & 60.4 \\ \hline
    %iteration 4 & 32.89 & & 32.5 & 49.3 & 69.7 \\ \hline
    \end{tabularx}
\end{table}

OurBT was very instrumental in helping us select the best dataset combinations as well as improving performance. The Luganda monolingual dataset combination that achieved the best BLEU score (35.94) for Eng2Lug (Table \ref{table:eng_to_lug_newtest} was Mozilla common voice, Makerere Text and Speech, and Youtube news headlines. The results further show that if we apply the standard BT on the same monolingual dataset combination, the performance only improves by 2.49 units. Our assumption is that when ourBT is applied, the model learns more contextual representation from each dataset that is used.

The English monolingual dataset combination that gave us the best BLEU score (39.97) for Lug2Eng model(Table \ref{table:lug_to_eng_newtest}) was Digital Umuganda, Gamayun MiniKit and Chimpreports news articles as seen in Table \ref{table:mono_english}. Similar to Eng2Lug, applying the StandardBT on the selected dataset only results in a 0.45 increase in the BLEU score. 

Even though we define no specific order in which these mini-datasets should be combined, our observation is that the ones that performed well had a similar contextual structure as our training dataset. Most importantly, almost 90\% of the data was collected from Uganda. We further observe the contribution of the monolingual web-crawled datasets as in both language directions, as they were part of the top dataset combinations. Another observation is that OurBT and IterBT resulted in a continuous decrease in the TER, which supplements the improvement in the BLEU score.   

\subsection{Sample translations}
We extract sample translations from our model and compare them with Google Translate. In both examples translated from English to Luganda (Table \ref{table:sample_eng2lug}), our model translates to the exact text as the reference. For Google Translate, it paraphrased the first example but it does not divert much from the meaning. However, for Example 2, it translates Maize (Kasooli) to coffee (Emmwaanyi). 

In Table \ref{table:sample_lug2eng}, the translation from Luganda to English shows that in Example 1, both our model and Google Translate paraphrase the sentences. However, Google Translate's use of the word "wasted" tends to differ a bit from the real meaning as per the Reference text. In example two, our model gives the correct translation compared to Google Translate which gives a different context.  

Generally, our observation indicates that our model results are highly comparable and sometimes better than those of Google Translate. 

  \begin{table*}[ht]
    \centering
    \caption{Sample Translations for Eng2Lug}
    \label{table:sample_eng2lug}
    %\begin{tabular}{|m{2cm}|m{3.5cm}|m{3.5cm}|m{3.5cm}|m{3.5cm}|}
    \begin{tabularx}{\textwidth}{|c|X|X|X|X|}
    %\begin{tabular}{|c|c|c||c|c|}
    \hline
    & \textbf{Source (Eng)} & \textbf{Target (Ours)} & \textbf{Google translate} & \textbf{Reference} \\ \hline
     1 & The chairperson of the company will address the farmers on the best farming practices & Ssentebe wa kampuni ajja kwogerako eri abalimi ku nnima ennungi & Ssentebe wa kampuni eno agenda kwogera eri abalimi ku nkola y’okulima ennungi & Ssentebe wa kampuni ajja kwogerako eri abalimi ku nnima ennungi \\ \hline
     2 & Maize need moderate rain to grow well & Kasooli yeetaaga enkuba ensaamuusaamu okukula obulungi & \textbf{\textit{Emmwaanyi}} zeetaaga enkuba ey’ekigero okusobola okukula obulungi & Kasooli yeetaaga enkuba ensaamuusaamu okukula obulungi \\ \hline
    \end{tabularx}
\end{table*}

\begin{table*}[ht]
    \centering
    \caption{Sample Translations for Lug2Eng}
    \label{table:sample_lug2eng}
    %\begin{tabular}{|m{2cm}|m{3.5cm}|m{3.5cm}|m{3.5cm}|m{3.5cm}|}
    \begin{tabularx}{\textwidth}{|c|X|X|X|X|}
    \hline
    & \textbf{Source (Lug)} & \textbf{Target (Ours)} & \textbf{Google translate} & \textbf{Reference} \\ \hline
     1 & Balowoza nti ssente zaabwe zibulankanyiziddwa & They think their money is embezzled & They think their money has been wasted & They assume that their money has been misused \\ \hline
     2 & Buli lusoma, ebibuuzo by'okwanukula bijja kuweebwa era bikozesebwe nga okulamula ensoma. & At every teaching , quiz questions will be given and used for performance evaluation & \textbf{\textit{Each semester, feedback}} questions will be given and used \textbf{\textit{as a judgment of the curriculum}}. &  At every teaching, quiz questions will be given and used for performance evaluation. \\ \hline
    \end{tabularx}
\end{table*}

\section{Discussion and conclusion}

This paper introduces benchmark results for English and Luganda translation pairs using BT, enhancing translation quality with bilingual and monolingual datasets, including custom datasets collected through web crawling. OurBT approach not only elevates BLEU scores but also optimizes monolingual dataset selection from independently sourced collections. Notably, we achieve BLEU scores of 40.25 (Eng2Lug) and 44.25 (Lug2Eng) with iterative BT (IteBT), surpassing existing studies that report lower BLEU improvements. In comparison, prior work achieved scores of 26.7 (Eng2Lug) and 33.2 (Lug2Eng) using BT, with a BLEU increment of approximately +3 \cite{akera2022machine}, while our approach boosts performance by +10 points. Beyond BLEU, we provide additional evaluation metrics, TER and ChrF to account for the agglutinative nature of Luganda. For example, while Google’s work \cite{bapna2022building} provides ChrF and notes BLEU’s limitations with morphologically rich languages, similar studies are sparse. We report TER improvements from 69.7 to 65.4 for Eng2Lug and from 64.7 to 60.4 for Lug2Eng, affirming our model’s advancing performance and emphasizing the importance of language-sensitive evaluation in low-resource machine translation \cite{babirye2022building}.

% This work has also contributed to reducing the data scarcity problem by providing sources and approaches that can be used for training NMT models for LRLs. We use web crawling techniques to access news headlines and articles that are in the public domain. Our understanding is that we respected the ethical boundaries in using this information. For instance, the full news articles were broken into sentences at every full stop, and then shuffled with the rest of the data. News articles are also admirable for their usage of professional language. 

% In general, NMT practitioners still grapple with defining the right ratios on how the real and synthetic parallel data can be applied in IncBT, even though a balance between the two is highly emphasized \cite{sennrich2015improving}. This issue was also studied by \cite{hoang2018iterative} where various ratios of the monolingual data were incrementally added showing that the more synthetic data, the better. However, it is not obvious on how the data can be incremented if collected from multiple sources. This is a major case for most LRLs where access to monolingual data is still limited with some languages having no digital presence, hence the alternative will still be to collect small datasets from multiple sources, merge them and perform BT or else build domain specific NMT models by selecting the most relevant sentences.

Balancing real and synthetic data in Incremental Back-Translation (IncBT) remains challenging, particularly for LRLs where monolingual data is scarce or fragmented across sources \cite{sennrich2015improving, hoang2018iterative}. For these languages, collecting small datasets from diverse sources and merging them for BT, or selecting domain-specific sentences, offers a viable approach to improve translation quality.

%Our research presents notable advancements in NMT for English-Luganda, targeting LRL challenges. By applying BT with custom NMT models and datasets—sourced from both public and web-crawled data—we surpassed existing benchmarks, achieving BLEU score improvements of over 10 points in both translation directions. These gains are validated through multiple evaluation metrics, including SacreBLEU, ChrF2, and TER, offering a comprehensive view of translation quality. Our novel dataset selection strategy further optimizes incremental BT’s effectiveness, providing a replicable framework for NMT in other LRLs.

Future work will explore multilingual methods, language models\cite{sujaini2023analysis} and unsupervised NMT, leveraging linguistic similarities among African languages, although limited access to diverse digital texts and LLM support for African languages, like Luganda, remains a challenge \cite{chauhan2022improved}. This research lays the groundwork for enhanced digital accessibility in African languages, fostering progress in low-resource machine translation.

\section*{Declarations}
The authors have no competing interests to declare relevant to this article's content.

%Authors should not prepare this section as a numbered or unnumbered {\verb|\section|}; please use the ``{\verb|acks|}'' environment.

%% The acknowledgments section 
\begin{acks}
This research was supported by Basic Science Research Program through the National Research Foundation of Korea funded by the Ministry of Education (NRF-2022R1A2C1012633), and the MSIT(Ministry of Science, ICT), Korea, under the Global Research Support Program in the Digital Field program(RS-2024-00431394) supervised by the IITP(Institute for Information \& Communications Technology Planning \& Evaluation).
\end{acks}

%%
%% We define the bibliography style to be used, 
%%\bibliographystyle{ieeetr}
\bibliographystyle{ACM-Reference-Format}
\bibliography{references.bib}


\begin{thebibliography}{37}

%%% the Web address.

\ifx \showCODEN    \undefined \def \showCODEN     #1{\unskip}     \fi
\ifx \showDOI      \undefined \def \showDOI       #1{#1}\fi
\ifx \showISBNx    \undefined \def \showISBNx     #1{\unskip}     \fi
\ifx \showISBNxiii \undefined \def \showISBNxiii  #1{\unskip}     \fi
\ifx \showISSN     \undefined \def \showISSN      #1{\unskip}     \fi
\ifx \showLCCN     \undefined \def \showLCCN      #1{\unskip}     \fi
\ifx \shownote     \undefined \def \shownote      #1{#1}          \fi
\ifx \showarticletitle \undefined \def \showarticletitle #1{#1}   \fi
\ifx \showURL      \undefined \def \showURL       {\relax}        \fi
% invisible to TeX
\providecommand\bibfield[2]{#2}
\providecommand\bibinfo[2]{#2}
\providecommand\natexlab[1]{#1}
\providecommand\showeprint[2][]{arXiv:#2}

\bibitem[Adelani et~al\mbox{.}(2023)]%
        {adelani2023masakhanews}
\bibfield{author}{\bibinfo{person}{David~Ifeoluwa Adelani}, \bibinfo{person}{Marek Masiak}, \bibinfo{person}{Israel~Abebe Azime}, \bibinfo{person}{Jesujoba~Oluwadara Alabi}, \bibinfo{person}{Atnafu~Lambebo Tonja}, \bibinfo{person}{Christine Mwase}, \bibinfo{person}{Odunayo Ogundepo}, \bibinfo{person}{Bonaventure~FP Dossou}, \bibinfo{person}{Akintunde Oladipo}, \bibinfo{person}{Doreen Nixdorf}, {et~al\mbox{.}}} \bibinfo{year}{2023}\natexlab{}.
\newblock \showarticletitle{Masakhanews: News topic classification for african languages}.
\newblock \bibinfo{journal}{\emph{arXiv preprint arXiv:2304.09972}} (\bibinfo{year}{2023}).
\newblock


\bibitem[Aji and Heafield(2020)]%
        {aji2020fully}
\bibfield{author}{\bibinfo{person}{Alham~Fikri Aji} {and} \bibinfo{person}{Kenneth Heafield}.} \bibinfo{year}{2020}\natexlab{}.
\newblock \showarticletitle{Fully synthetic data improves neural machine translation with knowledge distillation}.
\newblock \bibinfo{journal}{\emph{arXiv preprint arXiv:2012.15455}} (\bibinfo{year}{2020}).
\newblock


\bibitem[Akera et~al\mbox{.}(2022)]%
        {akera2022machine}
\bibfield{author}{\bibinfo{person}{Benjamin Akera}, \bibinfo{person}{Jonathan Mukiibi}, \bibinfo{person}{Lydia~Sanyu Naggayi}, \bibinfo{person}{Claire Babirye}, \bibinfo{person}{Isaac Owomugisha}, \bibinfo{person}{Solomon Nsumba}, \bibinfo{person}{Joyce Nakatumba-Nabende}, \bibinfo{person}{Engineer Bainomugisha}, \bibinfo{person}{Ernest Mwebaze}, {and} \bibinfo{person}{John Quinn}.} \bibinfo{year}{2022}\natexlab{}.
\newblock \showarticletitle{Machine translation for african languages: Community creation of datasets and models in uganda}. In \bibinfo{booktitle}{\emph{3rd Workshop on African Natural Language Processing}}.
\newblock


\bibitem[Anastasopoulos et~al\mbox{.}(2020)]%
        {anastasopoulos2020tico}
\bibfield{author}{\bibinfo{person}{Antonios Anastasopoulos}, \bibinfo{person}{Alessandro Cattelan}, \bibinfo{person}{Zi-Yi Dou}, \bibinfo{person}{Marcello Federico}, \bibinfo{person}{Christian Federman}, \bibinfo{person}{Dmitriy Genzel}, \bibinfo{person}{Francisco Guzm{\'a}n}, \bibinfo{person}{Junjie Hu}, \bibinfo{person}{Macduff Hughes}, \bibinfo{person}{Philipp Koehn}, {et~al\mbox{.}}} \bibinfo{year}{2020}\natexlab{}.
\newblock \showarticletitle{TICO-19: the translation initiative for COvid-19}.
\newblock \bibinfo{journal}{\emph{arXiv preprint arXiv:2007.01788}} (\bibinfo{year}{2020}).
\newblock


\bibitem[Babirye et~al\mbox{.}(2022)]%
        {babirye2022building}
\bibfield{author}{\bibinfo{person}{Claire Babirye}, \bibinfo{person}{Joyce Nakatumba-Nabende}, \bibinfo{person}{Andrew Katumba}, \bibinfo{person}{Ronald Ogwang}, \bibinfo{person}{Jeremy~Tusubira Francis}, \bibinfo{person}{Jonathan Mukiibi}, \bibinfo{person}{Medadi Ssentanda}, \bibinfo{person}{Lilian~D Wanzare}, {and} \bibinfo{person}{Davis David}.} \bibinfo{year}{2022}\natexlab{}.
\newblock \showarticletitle{Building text and speech datasets for low resourced languages: A case of languages in east Africa}.
\newblock  (\bibinfo{year}{2022}).
\newblock


\bibitem[Babirye et~al\mbox{.}(2023)]%
        {DVN/XSGIKR_2023}
\bibfield{author}{\bibinfo{person}{Claire Babirye}, \bibinfo{person}{Jeremy Tusubira}, \bibinfo{person}{Jonathan Mukiibi}, \bibinfo{person}{Joyce Nakatumba-Nabende}, {and} \bibinfo{person}{Andrew Katumba}.} \bibinfo{year}{2023}\natexlab{}.
\newblock \bibinfo{title}{{Sentiment Tagged Parallel Corpus for Luganda and Swahili}}.
\newblock
\newblock
\urldef\tempurl%
\url{https://doi.org/10.7910/DVN/XSGIKR}
\showDOI{\tempurl}


\bibitem[Baertlein and Ssekitto(2014)]%
        {baertlein2014luganda}
\bibfield{author}{\bibinfo{person}{Elizabeth Baertlein} {and} \bibinfo{person}{Martin Ssekitto}.} \bibinfo{year}{2014}\natexlab{}.
\newblock \showarticletitle{Luganda Nouns: Inflectional Morphology and Tests}.
\newblock \bibinfo{journal}{\emph{Linguistic Portfolios}} \bibinfo{volume}{3}, \bibinfo{number}{1} (\bibinfo{year}{2014}), \bibinfo{pages}{3}.
\newblock


\bibitem[Bapna et~al\mbox{.}(2022)]%
        {bapna2022building}
\bibfield{author}{\bibinfo{person}{Ankur Bapna}, \bibinfo{person}{Isaac Caswell}, \bibinfo{person}{Julia Kreutzer}, \bibinfo{person}{Orhan Firat}, \bibinfo{person}{Daan van Esch}, \bibinfo{person}{Aditya Siddhant}, \bibinfo{person}{Mengmeng Niu}, \bibinfo{person}{Pallavi Baljekar}, \bibinfo{person}{Xavier Garcia}, \bibinfo{person}{Wolfgang Macherey}, {et~al\mbox{.}}} \bibinfo{year}{2022}\natexlab{}.
\newblock \showarticletitle{Building machine translation systems for the next thousand languages}.
\newblock \bibinfo{journal}{\emph{arXiv preprint arXiv:2205.03983}} (\bibinfo{year}{2022}).
\newblock


\bibitem[Chauhan et~al\mbox{.}(2022)]%
        {chauhan2022improved}
\bibfield{author}{\bibinfo{person}{Shweta Chauhan}, \bibinfo{person}{Shefali Saxena}, {and} \bibinfo{person}{Philemon Daniel}.} \bibinfo{year}{2022}\natexlab{}.
\newblock \showarticletitle{Improved unsupervised neural machine translation with semantically weighted back translation for morphologically rich and low resource languages}.
\newblock \bibinfo{journal}{\emph{Neural Processing Letters}} \bibinfo{volume}{54}, \bibinfo{number}{3} (\bibinfo{year}{2022}), \bibinfo{pages}{1707--1726}.
\newblock


\bibitem[Cheng and Cheng(2019)]%
        {cheng2019joint}
\bibfield{author}{\bibinfo{person}{Yong Cheng} {and} \bibinfo{person}{Yong Cheng}.} \bibinfo{year}{2019}\natexlab{}.
\newblock \showarticletitle{Joint training for pivot-based neural machine translation}.
\newblock \bibinfo{journal}{\emph{Joint training for neural machine translation}} (\bibinfo{year}{2019}), \bibinfo{pages}{41--54}.
\newblock


\bibitem[Conneau et~al\mbox{.}(2019)]%
        {conneau2019unsupervised}
\bibfield{author}{\bibinfo{person}{Alexis Conneau}, \bibinfo{person}{Kartikay Khandelwal}, \bibinfo{person}{Naman Goyal}, \bibinfo{person}{Vishrav Chaudhary}, \bibinfo{person}{Guillaume Wenzek}, \bibinfo{person}{Francisco Guzm{\'a}n}, \bibinfo{person}{Edouard Grave}, \bibinfo{person}{Myle Ott}, \bibinfo{person}{Luke Zettlemoyer}, {and} \bibinfo{person}{Veselin Stoyanov}.} \bibinfo{year}{2019}\natexlab{}.
\newblock \showarticletitle{Unsupervised cross-lingual representation learning at scale}.
\newblock \bibinfo{journal}{\emph{arXiv preprint arXiv:1911.02116}} (\bibinfo{year}{2019}).
\newblock


\bibitem[Edunov et~al\mbox{.}(2018)]%
        {edunov2018understanding}
\bibfield{author}{\bibinfo{person}{Sergey Edunov}, \bibinfo{person}{Myle Ott}, \bibinfo{person}{Michael Auli}, {and} \bibinfo{person}{David Grangier}.} \bibinfo{year}{2018}\natexlab{}.
\newblock \showarticletitle{Understanding back-translation at scale}.
\newblock \bibinfo{journal}{\emph{arXiv preprint arXiv:1808.09381}} (\bibinfo{year}{2018}).
\newblock


\bibitem[Gatiatullin et~al\mbox{.}(2022)]%
        {gatiatullin2022toolset}
\bibfield{author}{\bibinfo{person}{Ayrat Gatiatullin}, \bibinfo{person}{Lenara Kubedinova}, \bibinfo{person}{Nikolai Prokopyev}, {and} \bibinfo{person}{Abduramanov Ibraim}.} \bibinfo{year}{2022}\natexlab{}.
\newblock \showarticletitle{Toolset of “Turkic Morpheme” Portal for Creation of Electronic Corpora of Turkic Languages in a Unified Conceptual Space}. In \bibinfo{booktitle}{\emph{2022 7th International Conference on Computer Science and Engineering (UBMK)}}. IEEE, \bibinfo{pages}{408--412}.
\newblock


\bibitem[Gitau and Marivate(2023)]%
        {gitau2023textual}
\bibfield{author}{\bibinfo{person}{Catherine Gitau} {and} \bibinfo{person}{Vukosi Marivate}.} \bibinfo{year}{2023}\natexlab{}.
\newblock \showarticletitle{Textual Augmentation Techniques Applied to Low Resource Machine Translation: Case of Swahili}.
\newblock \bibinfo{journal}{\emph{arXiv preprint arXiv:2306.07414}} (\bibinfo{year}{2023}).
\newblock


\bibitem[Gundecha(2015)]%
        {gundecha2015selenium}
\bibfield{author}{\bibinfo{person}{Unmesh Gundecha}.} \bibinfo{year}{2015}\natexlab{}.
\newblock \bibinfo{booktitle}{\emph{Selenium Testing Tools Cookbook}}.
\newblock \bibinfo{publisher}{Packt Publishing Ltd}.
\newblock


\bibitem[Hans and Milton(2016)]%
        {hans2016improving}
\bibfield{author}{\bibinfo{person}{Krupakar Hans} {and} \bibinfo{person}{RS Milton}.} \bibinfo{year}{2016}\natexlab{}.
\newblock \showarticletitle{Improving the performance of neural machine translation involving morphologically rich languages}.
\newblock \bibinfo{journal}{\emph{arXiv preprint arXiv:1612.02482}} (\bibinfo{year}{2016}).
\newblock


\bibitem[Hoang et~al\mbox{.}(2018)]%
        {hoang2018iterative}
\bibfield{author}{\bibinfo{person}{Cong Duy~Vu Hoang}, \bibinfo{person}{Philipp Koehn}, \bibinfo{person}{Gholamreza Haffari}, {and} \bibinfo{person}{Trevor Cohn}.} \bibinfo{year}{2018}\natexlab{}.
\newblock \showarticletitle{Iterative back-translation for neural machine translation}. In \bibinfo{booktitle}{\emph{2nd Workshop on Neural Machine Translation and Generation}}. Association for Computational Linguistics, \bibinfo{pages}{18--24}.
\newblock


\bibitem[Jais et~al\mbox{.}(2019)]%
        {jais2019adam}
\bibfield{author}{\bibinfo{person}{Imran Khan~Mohd Jais}, \bibinfo{person}{Amelia~Ritahani Ismail}, {and} \bibinfo{person}{Syed~Qamrun Nisa}.} \bibinfo{year}{2019}\natexlab{}.
\newblock \showarticletitle{Adam optimization algorithm for wide and deep neural network.}
\newblock \bibinfo{journal}{\emph{Knowl. Eng. Data Sci.}} \bibinfo{volume}{2}, \bibinfo{number}{1} (\bibinfo{year}{2019}), \bibinfo{pages}{41--46}.
\newblock


\bibitem[Jin et~al\mbox{.}(2022)]%
        {jin2022admix}
\bibfield{author}{\bibinfo{person}{Chang Jin}, \bibinfo{person}{Shigui Qiu}, \bibinfo{person}{Nini Xiao}, {and} \bibinfo{person}{Hao Jia}.} \bibinfo{year}{2022}\natexlab{}.
\newblock \showarticletitle{AdMix: A mixed sample data augmentation method for neural machine translation}.
\newblock \bibinfo{journal}{\emph{arXiv preprint arXiv:2205.04686}} (\bibinfo{year}{2022}).
\newblock


\bibitem[Khenglawt et~al\mbox{.}({[n.\,d.]})]%
        {khenglawtaddressing}
\bibfield{author}{\bibinfo{person}{Vanlalmuansangi Khenglawt}, \bibinfo{person}{Sahinur~Rahman Laskar}, \bibinfo{person}{Partha Pakray}, {and} \bibinfo{person}{Ajoy~Kumar Khan}.} \bibinfo{year}{[n.\,d.]}\natexlab{}.
\newblock \showarticletitle{Addressing data scarcity issue for English--Mizo neural machine translation using data augmentation and language model}.
\newblock \bibinfo{journal}{\emph{Journal of Intelligent \& Fuzzy Systems}} \bibinfo{number}{Preprint} (\bibinfo{year}{[n.\,d.]}), \bibinfo{pages}{1--11}.
\newblock


\bibitem[Kimera et~al\mbox{.}(2022)]%
        {kimera2022building}
\bibfield{author}{\bibinfo{person}{Richard Kimera}, \bibinfo{person}{Daniela~N Rim}, {and} \bibinfo{person}{Heeyoul Choi}.} \bibinfo{year}{2022}\natexlab{}.
\newblock \showarticletitle{Building a Parallel Corpus and Training Translation Models Between Luganda and English}.
\newblock \bibinfo{journal}{\emph{Information Science Journal}} \bibinfo{volume}{49}, \bibinfo{number}{11} (\bibinfo{year}{2022}), \bibinfo{pages}{1009--1016}.
\newblock


\bibitem[Lamar and Kaya(2023)]%
        {lamar2023measuring}
\bibfield{author}{\bibinfo{person}{Annie Lamar} {and} \bibinfo{person}{Zeyneb Kaya}.} \bibinfo{year}{2023}\natexlab{}.
\newblock \showarticletitle{Measuring the Impact of Data Augmentation Methods for Extremely Low-Resource NMT}. In \bibinfo{booktitle}{\emph{Proceedings of the The Sixth Workshop on Technologies for Machine Translation of Low-Resource Languages (LoResMT 2023)}}. \bibinfo{pages}{101--109}.
\newblock


\bibitem[Mojapelo and Buys(2023)]%
        {mojapelo2023data}
\bibfield{author}{\bibinfo{person}{Maxwell Mojapelo} {and} \bibinfo{person}{Jan Buys}.} \bibinfo{year}{2023}\natexlab{}.
\newblock \showarticletitle{Data augmentation for low resource neural machine translation for sotho-tswana languages}.
\newblock  (\bibinfo{year}{2023}).
\newblock


\bibitem[Mukiibi et~al\mbox{.}(2023)]%
        {DVN/EQOWTW_2023}
\bibfield{author}{\bibinfo{person}{Jonathan Mukiibi}, \bibinfo{person}{Claire Babirye}, \bibinfo{person}{Jeremy Tusubira}, \bibinfo{person}{Tobias Bateesa}, \bibinfo{person}{Eric~Peter Wairagala}, \bibinfo{person}{Chodrine Mutebi}, \bibinfo{person}{Joyce Nakatumba-Nabende}, \bibinfo{person}{Andrew Katumba}, \bibinfo{person}{Ivan Ssenkungu}, {and} \bibinfo{person}{Medadi Sentanda}.} \bibinfo{year}{2023}\natexlab{}.
\newblock \bibinfo{title}{{Luganda Monolingual Corpus}}.
\newblock
\newblock
\urldef\tempurl%
\url{https://doi.org/10.7910/DVN/EQOWTW}
\showDOI{\tempurl}


\bibitem[Mukiibi et~al\mbox{.}(2022)]%
        {mukiibi2022makerere}
\bibfield{author}{\bibinfo{person}{Jonathan Mukiibi}, \bibinfo{person}{Andrew Katumba}, \bibinfo{person}{Joyce Nakatumba-Nabende}, \bibinfo{person}{Ali Hussein}, {and} \bibinfo{person}{Josh Meyer}.} \bibinfo{year}{2022}\natexlab{}.
\newblock \showarticletitle{The makerere radio speech corpus: A Luganda radio corpus for automatic speech recognition}.
\newblock \bibinfo{journal}{\emph{arXiv preprint arXiv:2206.09790}} (\bibinfo{year}{2022}).
\newblock


\bibitem[Oh et~al\mbox{.}(2023)]%
        {oh2023data}
\bibfield{author}{\bibinfo{person}{Seokjin Oh}, \bibinfo{person}{Woohwan Jung}, {et~al\mbox{.}}} \bibinfo{year}{2023}\natexlab{}.
\newblock \showarticletitle{Data augmentation for neural machine translation using generative language model}.
\newblock \bibinfo{journal}{\emph{arXiv preprint arXiv:2307.16833}} (\bibinfo{year}{2023}).
\newblock


\bibitem[Rim et~al\mbox{.}(2023)]%
        {rim2023mini}
\bibfield{author}{\bibinfo{person}{Daniela~N Rim}, \bibinfo{person}{Richard Kimera}, {and} \bibinfo{person}{Heeyoul Choi}.} \bibinfo{year}{2023}\natexlab{}.
\newblock \showarticletitle{Mini-Batching with Similar-Length Sentences to Quickly Train NMT Models}.
\newblock \bibinfo{journal}{\emph{Information Science Journal}} \bibinfo{volume}{50}, \bibinfo{number}{7} (\bibinfo{year}{2023}), \bibinfo{pages}{614--620}.
\newblock


\bibitem[Sennrich et~al\mbox{.}(2015a)]%
        {sennrich2015improving}
\bibfield{author}{\bibinfo{person}{Rico Sennrich}, \bibinfo{person}{Barry Haddow}, {and} \bibinfo{person}{Alexandra Birch}.} \bibinfo{year}{2015}\natexlab{a}.
\newblock \showarticletitle{Improving neural machine translation models with monolingual data}.
\newblock \bibinfo{journal}{\emph{arXiv preprint arXiv:1511.06709}} (\bibinfo{year}{2015}).
\newblock


\bibitem[Sennrich et~al\mbox{.}(2015b)]%
        {sennrich2015neural}
\bibfield{author}{\bibinfo{person}{Rico Sennrich}, \bibinfo{person}{Barry Haddow}, {and} \bibinfo{person}{Alexandra Birch}.} \bibinfo{year}{2015}\natexlab{b}.
\newblock \showarticletitle{Neural machine translation of rare words with subword units}.
\newblock \bibinfo{journal}{\emph{arXiv preprint arXiv:1508.07909}} (\bibinfo{year}{2015}).
\newblock


\bibitem[Sennrich et~al\mbox{.}(2016)]%
        {sennrich2016neural}
\bibfield{author}{\bibinfo{person}{Rico Sennrich}, \bibinfo{person}{Barry Haddow}, {and} \bibinfo{person}{Alexandra Birch}.} \bibinfo{year}{2016}\natexlab{}.
\newblock \bibinfo{title}{Neural Machine Translation of Rare Words with Subword Units}.
\newblock
\newblock
\showeprint[arxiv]{1508.07909}~[cs.CL]


\bibitem[Ssentumbwe et~al\mbox{.}(2019)]%
        {ssentumbwe2019english}
\bibfield{author}{\bibinfo{person}{AM Ssentumbwe}, \bibinfo{person}{BM Kim}, {and} \bibinfo{person}{HA Lee}.} \bibinfo{year}{2019}\natexlab{}.
\newblock \showarticletitle{English to Luganda SMT: Ganda Noun Class Prefix Segmentation for Enriched Machine Translation [J]}.
\newblock \bibinfo{journal}{\emph{International Journal of Advanced Trends in Computer Science and Engineering}} \bibinfo{volume}{8}, \bibinfo{number}{5} (\bibinfo{year}{2019}), \bibinfo{pages}{1861--1868}.
\newblock


\bibitem[Sugiyama and Yoshinaga(2019)]%
        {sugiyama2019data}
\bibfield{author}{\bibinfo{person}{Amane Sugiyama} {and} \bibinfo{person}{Naoki Yoshinaga}.} \bibinfo{year}{2019}\natexlab{}.
\newblock \showarticletitle{Data augmentation using back-translation for context-aware neural machine translation}. In \bibinfo{booktitle}{\emph{Proceedings of the fourth workshop on discourse in machine translation (DiscoMT 2019)}}. \bibinfo{pages}{35--44}.
\newblock


\bibitem[Sujaini et~al\mbox{.}(2023)]%
        {sujaini2023analysis}
\bibfield{author}{\bibinfo{person}{Herry Sujaini}, \bibinfo{person}{Samuel Cahyawijaya}, {and} \bibinfo{person}{Arif~B Putra}.} \bibinfo{year}{2023}\natexlab{}.
\newblock \showarticletitle{Analysis of Language Model Role in Improving Machine Translation Accuracy for Extremely Low Resource Languages}.
\newblock \bibinfo{journal}{\emph{Journal of Advances in Information Technology}} \bibinfo{volume}{14}, \bibinfo{number}{5} (\bibinfo{year}{2023}).
\newblock


\bibitem[Sun et~al\mbox{.}(2021)]%
        {sun2021machine}
\bibfield{author}{\bibinfo{person}{Mengtao Sun}, \bibinfo{person}{Hao Wang}, \bibinfo{person}{Mark Pasquine}, {and} \bibinfo{person}{Ibrahim A.~Hameed}.} \bibinfo{year}{2021}\natexlab{}.
\newblock \showarticletitle{Machine translation in low-resource languages by an adversarial neural network}.
\newblock \bibinfo{journal}{\emph{Applied Sciences}} \bibinfo{volume}{11}, \bibinfo{number}{22} (\bibinfo{year}{2021}), \bibinfo{pages}{10860}.
\newblock


\bibitem[Tiedemann and Thottingal(2020)]%
        {tiedemann2020opus}
\bibfield{author}{\bibinfo{person}{J{\"o}rg Tiedemann} {and} \bibinfo{person}{Santhosh Thottingal}.} \bibinfo{year}{2020}\natexlab{}.
\newblock \showarticletitle{OPUS-MT--building open translation services for the world}. In \bibinfo{booktitle}{\emph{Proceedings of the 22nd Annual Conference of the European Association for Machine Translation}}. \bibinfo{pages}{479--480}.
\newblock


\bibitem[Vaswani et~al\mbox{.}(2017)]%
        {vaswani2017attention}
\bibfield{author}{\bibinfo{person}{Ashish Vaswani}, \bibinfo{person}{Noam Shazeer}, \bibinfo{person}{Niki Parmar}, \bibinfo{person}{Jakob Uszkoreit}, \bibinfo{person}{Llion Jones}, \bibinfo{person}{Aidan~N Gomez}, \bibinfo{person}{{\L}ukasz Kaiser}, {and} \bibinfo{person}{Illia Polosukhin}.} \bibinfo{year}{2017}\natexlab{}.
\newblock \showarticletitle{Attention is all you need}.
\newblock \bibinfo{journal}{\emph{Advances in neural information processing systems}}  \bibinfo{volume}{30} (\bibinfo{year}{2017}).
\newblock


\bibitem[Wenzek et~al\mbox{.}(2019)]%
        {wenzek2019ccnet}
\bibfield{author}{\bibinfo{person}{Guillaume Wenzek}, \bibinfo{person}{Marie-Anne Lachaux}, \bibinfo{person}{Alexis Conneau}, \bibinfo{person}{Vishrav Chaudhary}, \bibinfo{person}{Francisco Guzm{\'a}n}, \bibinfo{person}{Armand Joulin}, {and} \bibinfo{person}{Edouard Grave}.} \bibinfo{year}{2019}\natexlab{}.
\newblock \showarticletitle{CCNet: Extracting high quality monolingual datasets from web crawl data}.
\newblock \bibinfo{journal}{\emph{arXiv preprint arXiv:1911.00359}} (\bibinfo{year}{2019}).
\newblock


\end{thebibliography}

\end{document}